# Learning Why Things Change:
# The Difference-Based Causality Learner

**Mark Voortman**
*mark@voortman.name*
Decision Systems Laboratory
School of Information Sciences
University of Pittsburgh
Pittsburgh, PA, 15260, USA

**Denver Dash**[*]
*denver.h.dash@intel.com*
Intel Research
Pittsburgh, PA, 15213, USA

**Marek J. Druzdzel**[†]
*marek@sis.pitt.edu*
Decision Systems Laboratory
School of Information Sciences
University of Pittsburgh
Pittsburgh, PA, 15260, USA

## Abstract

In this paper, we present the *Difference-Based Causality Learner* (DBCL), an algorithm for learning a class of discrete-time dynamic models that represents all causation across time by means of difference equations driving change in a system. We motivate this representation with real-world mechanical systems and prove DBCL's correctness for learning structure from time series data, an endeavour that is complicated by the existence of latent derivatives that have to be detected. We also prove that, under common assumptions for causal discovery, DBCL will identify the presence or absence of feedback loops, making the model more useful for predicting the effects of manipulating variables when the system is in equilibrium. We argue analytically and show empirically the advantages of DBCL over vector autoregression (VAR) and Granger causality models as well as modified forms of Bayesian and constraint-based structure discovery algorithms. Finally, we show that our algorithm can discover causal directions of alpha rhythms in human brains from EEG data.

## 1 INTRODUCTION AND MOTIVATION

In the past 20 years in AI, the practice of learning causal models from data has received considerable attention [cf., Pearl and Verma, 1991, Cooper and Herskovits, 1992, Spirtes et al., 2000]. Existing methods are based on the formalism of structural equation models (SEMs), which originated in the econometrics literature over 50 years ago [cf., Strotz and Wold, 1960], and Bayesian networks [Pearl, 1987] which started the paradigm shift of graphical models in AI and machine learning 20 years ago. These methods have predominately focused on the learning of equilibrium (static) causal structure, and have recently gained inroads into mainstream scientific research, especially in biology [cf., Sachs et al., 2005].

Despite the success of these static methods, many real-world systems are dynamic in nature and are accurately modeled by systems of simultaneous differential equations. Temporal causality, in general, has been studied extensively in econometrics over the past four decades: Granger causality and vector autoregression (VAR) methods have become very influential [cf., Granger, 1969, Engle and Granger, 1987, Sims, 1980]. In AI, there has been work on learning Dynamic Bayesian Networks (DBNs) [Friedman et al., 1998] and modified Granger causality [Eichler and Didelez, 2007]. None of these models explicitly take into account the fact that many dynamic systems are based on differential equations. This makes their representations overly general for such systems, allowing arbitrary causal relations across time. In this paper, we show that differential equations impose strict constraints on cross-temporal causal edges, and we present a method that is capable of exploiting that fact.

This paper considers *Difference-Based Causal Models* (DBCMs), a class of discrete-time dynamic models inspired by Iwasaki and Simon [1994] that models all causation across time by means of difference equations driving change in the system. This paper presents the first method to learn DBCMs from data: the *Difference-Based Causality Learner* (DBCL). This algorithm treats differences as latent variables and conducts an efficient search to find them in the course of constructing a DBCM. This method exploits the fact that unknown derivatives have fixed relationships to known variables and so are easier to find than latent variables in general. We prove that DBCL correctly learns DBCMs given faithfulness and a conditional independence oracle, and show empirically that it is also robust in the sense of avoiding unnecessary calculations of higher-order derivatives, thus preventing mistakes due to numerical errors. We show that compared to Granger causality and VAR models, DBCL output is much more parsimonious and informative. We also show empirically that it outperforms variants of the PC algorithm and greedy Bayesian search algorithms that have been modified to assume DBCM structure. Finally, we prove that DBCL will always identify instantaneous feedback loops when the underlying system is a DBCM, making it easier to detect when an equilibrated model will be causal. To our knowledge, no other method for causal discovery is guaranteed to identify the presence or absence of feedback loops.

---

[*]Also Department of Biomedical Informatics, School of Medicine, University of Pittsburgh.

[†]Also Faculty of Computer Science, Białystok University of Technology, Wiejska 45A, 15-351 Białystok, Poland.

## 2 DIFFERENCE-BASED CAUSAL MODELS

Stated briefly, a DBCM is a discrete-time model, based on SEMs, with a graphical interpretation very similar to DBNs. Contemporaneous causation is allowed, i.e., like DBNs, variables can be caused by other variables in the same time slice. The defining characteristic of a DBCM is that all causation across time is due to a derivative (e.g., $\dot{x}$) causing a change in its integral (e.g., $x$). This cross-temporal restriction makes DBCMs a subset of causal models as defined by Pearl [2000] and structural equation models similar to those discussed 50 years ago by Strotz and Wold [1960]. DBCM-like models were discussed by Iwasaki and Simon [1994] and Dash [2003, 2005] to analyze causality in dynamic systems, but to date no algorithm exists to learn them from data.

As an example, consider the set of equations describing the motion of a damped simple harmonic oscillator (SHO). A block of mass $m$ is suspended from a spring in a viscous fluid and several different forces are acting on it, such as the forces resulting from the spring and that of gravity. The harmonic oscillator is an archetypal dynamic system, ubiquitous in nature. Although a linear system, it can form a good approximation to many nonlinear systems close to equilibrium. Furthermore, the $F = ma$ relationship is a canonical example of contemporaneous causation: applying a force to cause a body to accelerate instantly. Thus, although this system is simple, it illustrates many important points. Causal interactions even in such a simple system are problematic when using standard representations for causality, as we will show shortly.

Like all mechanical systems, the equations of motion for the harmonic oscillator are given by Newton's 2nd law describing the acceleration $a$ of the mass under the forces (due to the weight, due to the spring, $F_x$, and due to viscosity, $F_v$) acting on the block. These forces *instantaneously* determine $a$; furthermore, they indirectly determine the values of all integrals of $a$, in particular the velocity $v$ and the position $x$, of the block. The longer time passes, the more influence the forces have on those integrals. Writing this continuous time system as a discrete time model, $v$ and $x$ are approximately determined by the difference equations: $v^{t+1} = v^t + a^t \Delta t$ and $x^{t+1} = x^t + v^t \Delta t$, resulting in the *cross-temporal* causal links in the graph of Figure 1(a) and (b). *Thus, differential equation systems imply cross-temporal arcs with a regular structure.* DBCMs assume that all cross-temporal arcs are of this form.

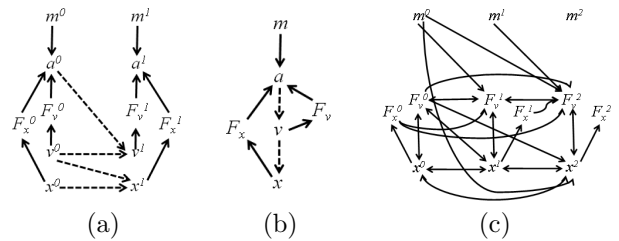

Figure 1: (a) The causal graph of a DBCM for the mass-spring system is always first-order Markovian. (b) The shorthand graph of (a) using dashed edges to indicate cross-temporal causation. (c) The unrolled graph with all $v$'s and $a$'s marginalized out is infinite-order Markovian.

More formally, DBCMs are a restricted form of structural equation models (SEMs). We first review these models, and then discuss our additional constraints. We use the notation $(A \perp\!\!\!\perp B \mid C)$ to indicate that variable $A$ is conditionally independent on $B$ given a set of variables $C$.

**Definition 1** (structural equation model). *A SEM is a pair $\langle V, E \rangle$, where $V = \{V_1, \ldots, V_n\}$ is a set of variables, and $E = \{E_1, \ldots, E_n\}$ is a set of equations such that each $E_i \in E$ can be written in the form: $V_i := f_i(W_i, \gamma_i)$ where $W_i \subseteq V \setminus V_i$ is called the set of causes (or parents) of $V_i$, denoted by $Pa(V_i)$, and the $\gamma_i$ are noise terms such that $(\gamma_i \perp\!\!\!\perp \gamma_j), i \neq j$.*

The *noise terms* $\gamma_i$ are intended to represent the set of causes of each variable that are not directly accounted for in the model. Historically, SEMs use linear equations with normally distributed noise terms.

A SEM defines a directed graph such that each variable $X \in V$ is represented by a node and there is an edge $Y \rightarrow X$ for each $Y \in Pa(X)$. In this way, SEMs can model relations between variables in a very general way. Furthermore, SEMs can be used to represent causality in dynamic systems for a discrete-time-setting by defining the set of variables to be a time-series: $\boldsymbol{V} = \boldsymbol{V^0} \cup \boldsymbol{V^1} \cup \boldsymbol{V^2}, \ldots$, where $\boldsymbol{V^t} = \{V_1^t, \ldots, V_n^t\}$ denotes the set of $n$ variables at time $t$. We call SEMs that partition their variables according to time indices *dynamic SEMs*.

DBCMs are a restricted form of dynamic SEMs. They assume that all causation across time is due to instantaneous causation of the *difference* of some variables:

**Definition 2** (Difference variable). *Let $\boldsymbol{V} = \boldsymbol{V^0} \cup \boldsymbol{V^1} \cup \boldsymbol{V^2} \cup \ldots$ be a time-series. The n-th order difference variable $\Delta^n V^t$ of variable $V^t \in \boldsymbol{V^t}$ is defined recursively as:*

$$\Delta^n V^t = \Delta^{n-1} V^{t+1} - \Delta^{n-1} V^t, \quad \text{with} \quad \Delta^0 V^t = V^t.$$

In particular: $\Delta^1 V^t = V^{t+1} - V^t$, which we sometimes shorten to $\Delta V^t$. When we invert the difference equation to give the value of $V^{t+1}$ in terms of its past value and its difference, we call it the *integral equation of $V^{t+1}$*. I.e., the integral equation of $V^{t+1}$ is $V^{t+1} = V^t + \Delta V^t$. Integral equations are identities and so are always deterministic. The graphs of Figure 1 use the standard notation from physics such that the derivative of $x$ is velocity ($v \equiv \Delta^1 x$) and the derivative of velocity is acceleration ($a \equiv \Delta^2 x$).

A DBCM is a dynamic SEM in which all causation across time is due to the presence of integral equations. Because all DBCMs are based on difference equations that do not vary from time to time, we can restrict ourselves to partitioning the variables into two time slices $\{0, 1\}$, where the 0th time slice determines the initial conditions and 1st time slice determines the transitions:

**Definition 3** (Difference-Based Causal Model). *A DBCM M is a dynamic SEM $S = \langle \boldsymbol{V}, \boldsymbol{E} \rangle$ with $\boldsymbol{V} = \boldsymbol{V^0} \cup \boldsymbol{V^1}$ and $\boldsymbol{E} = \boldsymbol{E^0} \cup \boldsymbol{E^1}$ such that there exists a cross-temporal parent of some variable $V_i^1 \in \boldsymbol{V^1}$ if and only if $E_i^1$ is the integral equation for variable $V_i^1$.*

This definition implies that the parent set of a variable $X^1$ that has parents in the previous time slice is $\boldsymbol{Pa}(X^1) = \{X^0, \Delta X^0\}$. If this is the case, we call $X$ an *integral* variable. An integral variable $X$ is part of a chain of causation $\Delta^j X^0 \to \Delta^{j-1} X^1 \to \ldots \to X^j$. We call the highest derivative ($\Delta^j X$) of this chain the *prime* variable of $X$, which we will also denote as $Prime(X)$. In the example of Figure 1, variables $x$ and $v$ are integral variables, and $a$ is the prime variable of $x$ and $v$.

Finally, any variable that is not an integral variable and is not a prime variable is called a *static* variable. This term does not imply that the variable is not changing from time-step to time-step, because it might have a causal ancestor that is part of an integration chain. However, we use this term to emphasize that the change is not due to a dynamic process involving these variables. In Figure 1, $m$, $F_v$ and $F_x$ are static variables.

The definition of DBCMs does not require that the contemporaneous structure be acyclic; however, in this paper we only consider acyclic DBCMs. It should be emphasized that this assumption does not restrict us to non-feedback systems; rather, this assumption implies that all feedback requires time to occur and thus will only occur through an integral variable. E.g., the position $x$ of the mass in the SHO causes an instantaneous spring force $F_x$ that results in an instantaneous acceleration $a$. Over time, $a$ causes a change in $x$ via integration. Thus we have the feedback loop: $x^0 \to F_x^0 \to a^0 \to v^1 \to x^2$. Although the instantaneous part, $x^0 \to F_x^0 \to a^0$, is acyclic, this is still a feedback system. $\boldsymbol{Fb}(X)$ is the set of instantaneous descendants of X which are also ancestors of $Prime(X)$ (in this example, $\boldsymbol{Fb}(x) = \{F_x\}$). Another interpretation is that by rejecting instantaneous loops we assume that the observation time-scale is much smaller than any time-scale of the system dynamics.

Since the contemporaneous structure is not changing over time, the equations in $\boldsymbol{E^0}$ and $\boldsymbol{E^1}$ are partially overlapping: those that correspond to contemporaneous structure are identical in both sets, but $\boldsymbol{E^0}$ contains initial conditions for all integral variables, and $\boldsymbol{E^1}$ contains integral equations for integral variables.

The graph in Figure 1(b) is a compressed version of the fully unrolled DBCM. The cycle in the graph caused by the dashed links is really an acyclic structure extending across time.

## 2.1 COMPARISON OF REPRESENTATIONS

Dynamic SEMs, like Granger causality and VAR models, allow arbitrary edges to exist across time. For many real physical systems this representation is too general.

DBCMs, by contrast, assume that all causation works in the same way as in mechanical systems. This restriction represents a tradeoff between expressibility and tractability. On one hand, DBCMs are only able to represent mechanical systems that are first-order Markovian. On the other hand, DBCMs are in principle easier to learn because, even when the derivatives are unobserved in the data as we will assume (e.g., in the previously introduced example we do not include $v$ and $a$, or any other derivative in the data set), at least we know something about these latent variables that are required to make the system Markovian.

When confronted with data that was generated by differential equations with some derivatives missing, the distinction between DBCL and the other approaches becomes glaring. Whereas, as we will show shortly, DBCL attempts to search for and identify the latent derivative variables, other approaches would try to marginalize them out. One might suspect that there is not much difference. For example, one might expect that a second order differential equation would simply result in a second-order Markov model when the derivatives are marginalized out. Unfortunately that is not the case, because the causation among the derivatives forms an infinite chain into the past. Thus, any approach that tries to marginalize out the derivatives must include infinite edges in the model, for example, such as those in Figure 1(c). In the harmonic oscillator system with all derivatives marginalized out, all parents of $a$ in time-slice $i$ of the DBCM are parents of $x$ for all time slices $j > i + 1$. Thus, the benefits of using the DBCM representation are not merely computational, but in fact, without learning the derivatives directly, the correct model does not have a finite representation.

## 3 DBCM LEARNING

The DBCM Learning problem can be posed in the following way: Given time-series data over a set of variables $\boldsymbol{V}$, derive a DBCM over a set of variables $\boldsymbol{V} \cup \boldsymbol{V_\Delta}$, where $\boldsymbol{V_\Delta}$ contains differences of variables that are derived from the original data. In other words, DBCL does not assume all relevant derivatives or the order of those derivatives are known. Instead, it treats these missing derivatives as latent variables and tries to discover them. We assume that, aside from these derivatives, there are no other latent confounding variables present. Note, for example, that this assumption also rules out the existence of structures of the form $\Delta X \to X \to Y$, where $\Delta X$ and $X$ are latent and $Y$ is observable, because the $X$ process forms a latent chain across time that can confound $Y$ at different times.

DBCL relies on the standard assumption of *faithfulness* [Spirtes et al., 2000]. Faithfulness is the converse of the Markov condition, and it is the critical assumption that allows structure to be uncovered from independence relations. However, when a dynamic system goes through equilibrium, *by definition*, faithfulness is violated. For example, if the motion of the block in the simple harmonic oscillator reaches equilibrium then, by definition, the equation $a = (F_x + F_v + mg)/m$ becomes $0 = F_x + F_v + mg$. This means that the values of the forces acting on the block are no longer correlated with the value of $a$, even though they are direct causes of $a$. Thus, by assuming faithfulness, we are implicitly assuming that no equilibrations have occurred.

### 3.1 THE ALGORITHM

DBCL consists of two steps: (1) detecting prime (and integral) variables ($\boldsymbol{V_\Delta}$), and (2) learning the contemporaneous structure. The first step is achieved by calculating numerical derivatives of all variables and then deciding which ones should be prime variables. This is based on the following theorem, which exploits the fact that only prime variables can always be made independent across time by conditioning on variables in $\boldsymbol{V^1}$:

**Theorem 1** (detecting prime variables). *Let $I$ be the set of conditional independence relations implied by faithfulness applied to a DBCM $M = \langle \boldsymbol{V}, \boldsymbol{E} \rangle$ with $\boldsymbol{V} = \boldsymbol{V^0} \cup \boldsymbol{V^1}$ and $\boldsymbol{E} = \boldsymbol{E^0} \cup \boldsymbol{E^1}$. Let $\Delta^j X^0$ denote the j-th order difference of some $X^0 \in \boldsymbol{V^0}$. Then $\Delta^j X^0$ is the prime variable of $X^0$ if and only if it can be d-separated from itself in the future and none of the lower order differences can be d-separated, i.e.:*

1. *there exists a $\boldsymbol{W} \subset \boldsymbol{V^1}$ such that $(\Delta^j X^0 \perp\!\!\!\perp \Delta^j X^1 \mid \boldsymbol{W}) \in I$, and*

2. *there exists no set $\boldsymbol{W'} \subset \boldsymbol{V^1}$ such that $(\Delta^k X^0 \perp\!\!\!\perp \Delta^k X^1 \mid \boldsymbol{W'}) \in I$, for all $k < j$.*

Once we have found $\boldsymbol{V_\Delta}$, the set of integral and prime variables in the model, learning contemporaneous structure over the two-time-slice model becomes a problem of learning a time-series model from *causally sufficient* data (i.e., there do not exist any latent common causes).

Theorem 2 shows that we can learn the contemporaneous structure from time-series data despite the fact that data from time-to-time is not independent. This is because we know by construction that the integral variables will d-separate the time slices, so all structure between variables can be obtained by conditioning only on variables in the same time slice:

**Theorem 2** (learning contemporaneous structure). *Let $I$ be the set of conditional independence relations implied by faithfulness applied to a DBCM $M = \langle \boldsymbol{V}, \boldsymbol{E} \rangle$, where $\boldsymbol{V} = \boldsymbol{V^0} \cup \boldsymbol{V^1}$. There is an edge $X^1 - Y^1$ if and only if:*

1. *Either $X^1$ or $Y^1$ is not an integral variable, and*

2. *there exists no $\boldsymbol{V'^1} \subset \boldsymbol{V^1} \setminus \{X^1, Y^1\}$ such that $(X^1 \perp\!\!\!\perp Y^1 \mid \boldsymbol{V'^1}) \in I$.*

In addition to having discovered the latent variables in the data, and the structure between non-integral variables, we also know that there can be no contemporaneous edges between two integral variables, and integral variables can have only outgoing contemporaneous edges. We can thus restrict the search space of causal structures. Theorems 1 and 2 together with these constraints form the basis of the DBCL algorithm:

**Algorithm 1** (DBCL (sketch)).
 **Input:** *a maximum difference parameter $k_{max} > 0$, a time-series dataset D over a set of variables $\boldsymbol{V'} = \boldsymbol{V^{0'}} \cup \boldsymbol{V^{1'}}$.*
**Output:** *a set $\boldsymbol{V_\Delta}$ of prime and integral variables, and a partially directed graph G over $\boldsymbol{V} = \boldsymbol{V'} \cup \boldsymbol{V_\Delta}$.*

1. *Find relevant latent derivatives (Theorem 1):*
   (a) *Initialize $k = 0$, and let $\boldsymbol{V_\Delta}$ be all differences up to $k_{max}$.*
   (b) *Let $\boldsymbol{W}$ be all variables plus their differences up to kth order that are in $\boldsymbol{V_\Delta}$.*
   (c) *For all $V \in \boldsymbol{V'}$ without a prime variable, check to see if there exists a set $\boldsymbol{W'} \subset \boldsymbol{W}$ that renders $\Delta^i V^0$ independent of $\Delta^i V^1$, for the $i \leq k$. If so, remove all $\Delta^j V^1$, $j > i'$, from $\boldsymbol{V_\Delta}$ where $i'$ is the lowest $i$ for which the independence occurred.*
   (d) *Let $k = k + 1$. If not all prime variables have been found and $k \leq k_{max}$, go to Step 1b.*

2. *Learn the structure (Theorem 2):*
   (a) *Learn the contemporaneous structure by using any correct causal discovery algorithm under causally sufficient data. Impose the following constraints:*
      i. *Forbid edges between all integral variables.*
      ii. *If X is an integral variable with an edge $X - Y$, direct the edge such that $X \to Y$.*
   (b) *Add all cross-temporal links specified by the set of integral and prime variables.*

The output of DBCL will depend on the algorithm used in Step 2. Our implementation is based on the constraint-based search PC algorithm, so the contemporaneous structure will be a partially directed graph that represents the statistical equivalence class in which the true directed graph belongs. One might argue that because there are deterministic relationships (the integral equations) in a DBCM, the faithfulness assumption is not valid. However, all deterministic relations involve exactly 3 variables, e.g., $\Delta X^0$, $X^0$ and $X^1$. However two of those variables are in time slice 0, so DBCL's conditioning tests never involve all three variables at the same time. The hidden variable thus effectively adds noise to the deterministic relationship.

## 3.2 IDENTIFICATION OF EMC VIOLATION

Given a model output from DBCL, one might be interested in performing causal inference, i.e., predicting the effects of manipulating components of the system. This operation is complicated by the presence of equilibrations that may have occurred in the system. Dash [2003, 2005] shows that some dynamic systems do not obey *Equilibration-Manipulation Commutability* (EMC), i.e., the causal graph that results when an equilibrium model is manipulated can be different from the (true) graph that results when the dynamic model is manipulated and then equilibrated. Dash points out two conditions which aid in EMC identification: First, if a variable is *self-regulating*, meaning that $X \in Pa(Prime(X))$, then when $X$ is equilibrated, the parent set of $X$ and the children set of $X$ are unchanged. Thus, with respect to manipulations on $X$, the EMC condition is obeyed. Second, a sufficient condition for the violation of EMC exists when the set of feedback variables of some (non-self-regulating) $X$ is nonempty in the equilibrium graph. In this case, there will always exist a manipulation that violates EMC.

Given a DBCM with all edges oriented, it is trivial to check these two conditions; however, since DBCL is not guaranteed to find the orientation of every edge in the DBCM structure, it is not obvious that DBCL is useful for identifying EMC violation. The following theorem shows that DBCL output will always identify self-regulating variables:

**Theorem 3.** *Let D be a DBCM with a variable X that has a prime variable $Prime(X)$. The partially directed graph returned by Algorithm 1 with a perfect independence oracle will have an edge between X and $Prime(X)$ if and only if X is self-regulating.*

It is easy to show that a feedback set of $X$ is empty if and only if all paths from $X$ to $Prime(X)$ have a collider. Again, since DBCL is not guaranteed to identify all edge orientations, not all colliders are necessarily identified. According to the faithfulness condition, DBCL will detect a correct equivalence class, and so will detect the correct adjacencies and the correct v-structures (unshielded colliders); thus Theorem 4 shows that we can always identify whether or not $\boldsymbol{Fb}(X)$ is empty:

**Theorem 4.** *Let G be the contemporaneous graph of a DBC model. Then for a variable X in G, $\boldsymbol{Fb}(X) = \emptyset$ if and only if for each undirected path $P = \langle P_0, P_1, \ldots, P_n \rangle$ between $P_0 = X$ and $P_n = Prime(X)$, there exists a v-structure $P_i \to P_j \leftarrow P_k$ in G such that $P_i, P_j, P_k \in P$.*

Theorem 4 asserts that we can determine whether or not there exists a directed path from $X$ to $Prime(X)$. In fact, this theorem does not make use of the fact that the path terminates on a prime variable, so it actually serves as an identifiability proof for all causal descendants of any integral variable.

## 4 RESULTS

For our empirical studies, we generated data from real physical systems that are representative of the type of systems found in nature. We also applied DBCL to real EEG brain data to reveal the causal propagation of alpha waves.[1]

Validation of DBCL is complicated by the fact that, as far as we know, there exist few suitable baseline methods that are even in principle able to correctly learn a DBCM when derivatives are unknown. As discussed in Section 2.1, if one tries to learn causal relations with the latent variables marginalized out, an infinite-order Markov model results (Figure 1(c)). The FCI algorithm [Spirtes et al., 2000], which attempts to take into consideration latent variables, would also result in an infinite-order Markov model because it does not try to isolate and learn the latent variables and the structure between them and the observables. The structural EM algorithm [Friedman, 1998] does try to learn explicit latent variables and structure. However, applying it naively would be unfair since DBCL uses background information about the latent variables that structural EM would not be privy to.

Thus, in order to provide a fair baseline, we chose to adapt some standard algorithms for discovery from causally sufficient data by providing them with known information about the latent derivatives. We used both the PC algorithm and a greedy Bayesian approach on a data set with all differences up to some maximum $k_{max} = 3$ calculated a priori, and we applied some heuristics to interpret the output as a DBCM. While perhaps not fully satisfying, we felt that this provided the fairest comparison to a baseline. Essentially, this allows us to assess how well Step 1 of DBCL (the main novel component) performed on learning latent differences. Once those latent differences are found, we used the PC algorithm and the Bayesian search algorithm to recover the contemporaneous structure, but without imposing the structure of a DBCM. In PC and DBCM we used a significance level of 0.01. The Bayesian approach starts with an empty network and then first greedily adds arcs using a Bayesian score with the K2 prior [Cooper and Herskovits, 1992], and then greedily removes arcs. For the Bayesian approach we discretized the data into five bins with approximately equal counts. It is possible to use a Bayesian approach without discretizing the data [Geiger and Heckerman, 1994], which we may explore in the future.

### 4.1 HARMONIC OSCILLATORS

We tested DBCL on models of two physical systems, namely a SHO and the more complex coupled SHO shown in Figure 2(a). Although the SHO is a deter-ministic system, having noise is still realistic: e.g., friction, air pressure, temperature, all of these factors are weak latent causes that add noise when determining the forces of the system. Thus all non-integral equations used Gaussian error terms that were resampled at every time interval. For both systems we selected parameters of our models in such a way that they were stable, i.e., produced measurements within reasonable bounds. We generated 100 data sets of 5,000 records for each system. We should emphasize that, as mentioned earlier, we do not include the derivatives in the data set, but only the original variables.

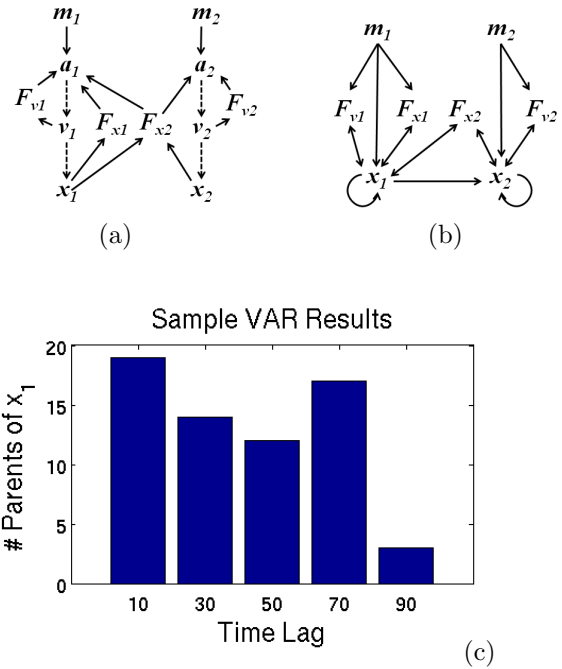

Figure 2: (a) The causal graph of the coupled SHO system. (b) A typical Granger causality graph recovered with simulated data. (c) The number of parents of $x_1$ over time-lag recovered from a VAR model (typical results).

We first computed Granger causality models and VAR models for some of the simulated data for the coupled SHO just to illustrate how uninformative these models are when the latent derivatives are unknown. Those results are shown in Figure 2(b) and Figure 2(c), respectively. The Granger graph is more difficult to interpret than the DBCM because of the presence of multiple double-headed edges indicating latent confounders. It was noted that the sole integral variables appeared in the Granger graph with reflexive edges, which might lead to an alternative algorithm for finding prime variables. However, the Granger graph does not provide enough information to perform causal reasoning. The VAR model is also difficult to interpret, as it attempts to learn structure over time of an infinite-order Markov model. The graph of Figure 2(c) shows that variable $x_1$ has 65 parents spread out over time-lags from 1 to 100 (binned into groups of $\Delta t = 20$) at significance level of 0.05. Thus while VAR models might be useful for prediction, they provide little insight into the causality of DBCMs.

---

[1] All experiments were performed with SMILE, a Bayesian inference engine developed at the Decision Systems Laboratory and available at http://genie.sis.pitt.edu/.

There were four algorithms used for quantitative baselines: two based on the PC algorithm and two based on the Bayesian algorithm. We will call them $PC_1$, $PC_2$, $B_1$, and $B_2$, respectively. For all baselines the procedure for detecting the prime variables was the same: all derivatives up to a maximum order were precalculated, and prime variables were determined to be the lowest order derivative that was not connected to itself in the future in the output graph. In the second step $PC_1$ and $B_1$ reported the contemporaneous structure that was found during the search for prime variables. For $PC_2$ and $B_2$, a separate step was made wherein we created a new dataset using only the derivatives found in step 1, and relearned contemporaneous structure from this reduced dataset. The results for the SHO are shown in the following table:

|        | $\%\Delta_{low}$ | $\%\Delta_{hi}$ | $\%E_{del}$ | $\%E_{add}$ | $\%O_{err}$ |
|--------|------|------|------|------|------|
| $PC_1$ | 0.00 | 0.50 | 39   | 230  | 26   |
| $PC_2$ | 0.00 | 0.50 | 100  | 100  | 1.0  |
| $B_1$  | 17   | 72   | 60   | 200  | 20   |
| $B_2$  | 17   | 72   | 78   | 120  | 14   |
| $DBC$  | 0.00 | 0.50 | 0.40 | 1.2  | 0.60 |

The first two columns of the table show the percentage of derivatives too low and to high, respectively. The other three columns of the table show the percentage of edges that were deleted, added, and incorrectly oriented. For example, on average, DBCL added 1.2 extra edges for every 100 edges in the correct graph, whereas $PC_2$ added 104 extra edges per 100 original edges.

The table below shows the results for the coupled SHO:

|        | $\%\Delta_{low}$ | $\%\Delta_{hi}$ | $\%E_{del}$ | $\%E_{add}$ | $\%O_{err}$ |
|--------|------|------|------|------|------|
| $PC_1$ | 0.00 | 12   | 40   | 200  | 23   |
| $PC_2$ | 0.00 | 12   | 84   | 26   | 14   |
| $B_1$  | 0.00 | 93   | 64   | 170  | 8.5  |
| $B_2$  | 0.00 | 93   | 42   | 140  | 21   |
| $DBC$  | 0.00 | 0.25 | 0.58 | 1.3  | 6.4  |

These results show that DBCL is effective at both learning the correct difference variables and of learning contemporaneous structure of these systems. For the SHO, the PC baselines are performing as well as DBCL for discovering prime variables; however, when the network gets more complicated, there is a clear difference. Also, in all cases the second step makes a big difference between baselines and DBCM, most likely because enforcing the DBCM structure is essential. We did try other significance levels besides 0.01, but all results showed the same trend.

## 4.2 EEG BRAIN DATA

In our second experiment, we attempted to learn a DBCM of the causal propagation of alpha waves, an 8-12 Hz signal that typically occurs in the human brain when the subject is in a waking state with eyes closed. Subjects were asked to close their eyes and then an EEG measurement was recorded. The data consisted of 10 subjects and a multivariate time series of 19 variables was recorded for each subject [2], containing over 200,000 time steps at a sampling rate of 256 Hz. Each variable corresponds to a brain region using the standard 10-20 convention for placement of the electrodes on the human scalp.

Alpha rhythms are known to operate in a specific frequency band peaking at 10 Hz. To focus our results more on this process, we tried learning a DBCM using just the 10 Hz power signal over time. We divided the data into 0.5s segments, performed a FFT on that segment and extracted the power of the 10 Hz bin for each time slice. When learning the DBCM, we used the same significance level and $k_{max}$ as before. The result for subject 10 is displayed in Figure 3. The circles represent the 19 variables that correspond to the brain regions. The top of this graph represents the front of the brain and the bottom the back. The small squares in each circle represent the derivatives that were found. The lower left is the original EEG signal, the lower right the first derivative, the top right the second derivative, and the top left the third derivative. In some regions, no derivatives were detected, so those squares have been left out.

Here (and in typical subjects) there are only a few regions that required derivatives to explain their variation. The locations of those regions varied quite a bit from subject to subject, but there were some common patterns. Across all subjects, 16 of 20 occipital regions had at least one derivative present. This contrasts to the frontal lobes where across all subjects only 1 of 70 frontal regions had one derivative or more. When a region had at least one derivative, rarely, if ever, did it also have an incoming edge from some region that did not have a derivative. This suggests that the regions containing the dynamic processes were the primary drivers of alpha-wave activity. Since most of these drivers occurred in the occipital lobes, this is consistent with the widely accepted view that alpha waves originate from the visual cortex.

---

[2]Data available at http://www.causality.inf.ethz.ch /repository.php?id=17

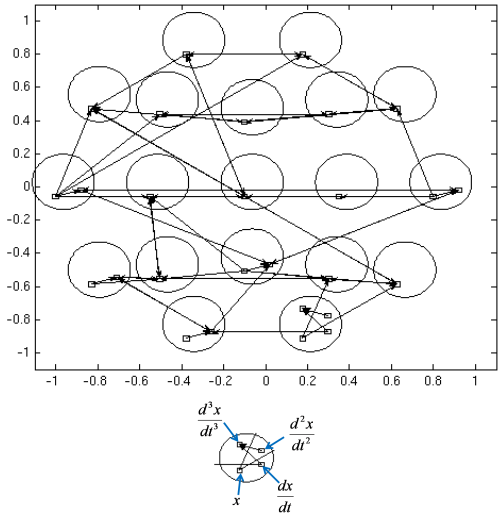

Figure 3: Output of DBCL on data filtered for alpha wave power.

There were many regions that did not require any derivatives to explain their signals. The alpha wave activity in these regions is quickly ($< 0.5s$) determined given the state of the generating regions. One hypothesis to explain this is given by Gòmez-Herrero et al. [2008] where they point out that conductivity of the skull can have significant impact on EEG readings by causing local signals to be a superposition of readings across the brain. Thus, if the readings of alpha waves detected in, say, the frontal region of the brain is due merely to conductivity of the skull, we would have effectively instantaneous determination of the alpha signal in those regions given the value in the regions generating the alpha waves.

We should note that when DBCL is applied to the raw (unfiltered) data, the resulting DBCM is much less neat: Most regions have at least one derivative present, and connectivity among regions is much higher and more difficult to interpret. This is not surprising given the massively parallel activity occuring in the brain, and it suggests that when seeking to learn causal interactions in the brain, it may be useful to partition brain signals into different frequency bands. We hope to look more fully into different bands and possibly for causal interactions among different bands.

## 5  DISCUSSION

The main contribution of this work is to present a new representation for learning models from time-series data. While DBCMs have been discussed elsewhere in terms of analyzing causality of dynamic systems, there has not as yet been an algorithm to learn them from data. This paper presents such an algorithm and, in the process, makes DBCMs accessible to a wide range of practitioners in econometrics, biology and AI who currently rely on Granger causality, vector autoregression or graphical models to model dynamic systems. We have argued that DBCMs are particularly suited to learning systems which are based on differential equations, and have shown empirically that, for such systems when the relevant derivatives are unknown, DBCL will learn models accurately where existing approaches will fail. We have proven that under common assumptions DBCL will learn the correct equivalence class for a DBCM, and have shown that several important feautures of the underlying DBCM are identifiable from this equivalence class, such as the presence of feedback loops and the set of descendants of integral variables.

While there exist mathematical dynamic systems that cannot be written as DBCMs, we believe that systems based on differential equations are ubiquitous in nature, and, therefore, will be well approximated by DBCMs. Furthermore, we have argued that there does not exist a representation that is capable of learning a finite model of these systems without first finding the correct latent derivative variables. This is because marginalizing out latent derivative variables results in an infinite-order Markov model. Thus our method can be viewed as contributing to the very hard problem of discovering latent common causes in difference equation systems.

We have also shown that DBCL can learn parsimonious representations for causal interactions of alpha waves in human brains that are consistent with previous research. We plan to apply this method to understanding causal pathways in the brain more broadly using a combination of EEG and MEG brain data.

In general, we find it surprising that after nearly 50 years of developing theories for identification of causes in econometrics, rarely, if ever, have researchers attempted to apply these theories to even the simplest dynamic physical systems. We feel that our work thus exposes a glaring gap in causal discovery and representation, and we hope that by reversing that process—applying a representation that works well on known mechanical systems to more complicated biological, econometric and AI systems—we can make new inroads to causal understanding in these disciplines.

## Acknowledgments

Marek Druzdzel was supported in part by the National Institute of Health under grant number U01HL101066-01. We would like to the thank the anonymous reviewers for their useful feedback.

## APPENDIX: PROOFS

**Theorem 1** (detecting prime variables). *Let $I$ be the set of conditional independence relations implied by faithfulness applied to a DBCM $M = \langle \boldsymbol{V}, \boldsymbol{E} \rangle$ with $\boldsymbol{V} = \boldsymbol{V^0} \cup \boldsymbol{V^1}$ and $\boldsymbol{E} = \boldsymbol{E^0} \cup \boldsymbol{E^1}$. Let $\Delta^j X^1$ denote the j-th order difference of some $X^1 \in \boldsymbol{V^1}$. Then $\Delta^j X^1$ is the prime variable of $X^1$ if and only if it is d-separated from itself in the future and none of the lower order differences can be d-separated, i.e.:*

1. *there exists a $\boldsymbol{W} \subset \boldsymbol{V^1}$ such that $(\Delta^j X^0 \perp\!\!\!\perp \Delta^j X^1 \mid \boldsymbol{W}) \in I$, and*

2. *there exists no set $\boldsymbol{W'} \subset \boldsymbol{V^1}$ such that $(\Delta^k X^0 \perp\!\!\!\perp \Delta^k X^1 \mid \boldsymbol{W'}) \in I$, for all $k < j$.*

*Proof.* The Markov and faithfulness conditions together imply that an edge exists between any two variables $A$ and $B$ in the model if and only if there exists no set $\boldsymbol{W}$ such that $(A \perp\!\!\!\perp B \mid \boldsymbol{W})$. In a DBCM model, an edge exists across time slices from a difference variable $A^0 \to A^1$ iff $A$ is an integral variable. Thus, the first difference variable for which there is no edge must be the prime variable, and Conditions 1 and 2 follow from the Markov and Faithfulness conditions. $\square$

**Theorem 2** (learning contemporaneous structure). *Let $I$ be the set of conditional independence relations implied by faithfulness applied to a DBCM $M = \langle \boldsymbol{V}, \boldsymbol{E} \rangle$, where $\boldsymbol{V} = \boldsymbol{V^0} \cup \boldsymbol{V^1}$. There is an edge $X^1 - Y^1$ if and only if:*

1. *Either $X^1$ or $Y^1$ is not an integral variable, and*

2. *there exists no $\boldsymbol{V'^1} \subset \boldsymbol{V^1} \setminus \{X^1, Y^1\}$ such that $(X^1 \perp\!\!\!\perp Y^1 \mid \boldsymbol{V'^1}) \in I$.*

*Proof.* ⇒ Condition 1 follows from the definition of DBCMs: an integral variable is never connected to another integral variable in a given time slice. Condition 2 follows from the Faithfulness condition. Faithfulness states that an independence relation implies a d-separation in the graph. Therefore the contrapositive states that a conection in the graph implies no conditional independence relation exists when conditioning over any subset of $\boldsymbol{V} \setminus \{X^1, Y^1\}$.

⇐ Assume there exists no conditional independence relation $(X^1 \perp\!\!\!\perp Y^1 \mid \boldsymbol{V'}) \in I$. We prove by contradiction that there must be an edge $X^1 - Y^1$. Assume there is no such edge. Then by the Markov condition there exists some set $\boldsymbol{V'} \subset \boldsymbol{V}$ such that $(X^1 \perp\!\!\!\perp Y^1 \mid \boldsymbol{V'}) \in I$. Let $\boldsymbol{V}_\Delta^1$ be the set of integral variables in time slice 1. According to the Markov condition, conditioning on these variables renders $\boldsymbol{V^0}$ independent of $\boldsymbol{V^1}$. Thus, the set $\{\boldsymbol{V'} \cup \boldsymbol{V}_\Delta^1\} \cap \{\boldsymbol{V^1} \setminus \{X^1, Y^1\}\} \subset \boldsymbol{V^1}$ will also render $X^1$ independent from $Y^1$, which contradicts our original assumption. Therefore, there must be an edge $X^1 - Y^1$. □

**Theorem 3.** *Let D be a DBCM with a variable X that has a prime variable $Prime(X)$. The partially directed graph returned by Algorithm 1 with a perfect independence oracle will have an edge between X and $Prime(X)$ if and only if X is self-regulating.*

*Proof.* Follows by the correctness of the structure discovery algorithm (all adjacencies in the graph will be recovered) together with the definition of DBCMs (no contemporaneous edge can be oriented into an integral variable). □

**Theorem 4.** *Let G be the contemporaneous graph of a DBC model. Then for a variable X in G, $\boldsymbol{Fb}(X) = \emptyset$ if and only if for each undirected path P between X and $Prime(X)$, there exists a v-structure $P_i \rightarrow P_j \leftarrow P_k$ in G such that $\{P_i, P_j, P_k\} \subset P$.*

*Proof.* ⇒ Assume $\boldsymbol{Fb}(X) = \emptyset$. Let P be an arbitrary path $P = P_0 \rightarrow P_1 - P_2 - \ldots - P_n - P_{n+1}$ with $P_0 = X$ and $P_{n+1} = Prime(X)$, and let k be the number of cross-path colliders on that path. The path must have at least one (cross-path) collider, otherwise there will be a directed path from X to $Prime(X)$ which contradicts the fact that $\boldsymbol{Fb}(X) = \emptyset$. If at least one of the cross-path colliders is unshielded the theorem is satisfied, so we only have to consider the case of shielded colliders. Now let $P_i \rightarrow P_j \leftarrow P_k$ be the first shielded cross-path collider (such that $j$ is the smallest). We consider three cases:

1. $i < j < k$: There is a directed path from X to $P_i$ since it is the first collider. Therefore, there can be no edge from $P_k$ to $P_i$, because that would create a collider in $P_i$ (and $P_j$ would not be the first). So there must be an edge from $P_i$ to $P_k$ and this implies there is a directed path from X to $P_k$ and we recurse and look for the first shielded cross-path collider after $P_k$.

2. $i, k < j$: Without loss of generality, there is a path $X \rightarrow \ldots \rightarrow P_i \rightarrow \ldots \rightarrow P_k \rightarrow \ldots \rightarrow P_j$, and edges $P_i \rightarrow P_j$, $P_k \rightarrow P_j$, and $P_i - P_k$. If $P_i \leftarrow P_k$ then there would be a collider in $P_i$ which contradicts that $P_j$ is the first one. Therefore, there must be an edge $P_i \rightarrow P_k$ and this implies there is a directed path from X to $P_j$ and we recurse and find the first shielded cross-path collider after $P_j$.

3. $j < i, k$: Without loss of generality, there is a path $X \rightarrow \ldots \rightarrow P_j \ldots P_i \ldots P_k$, and edges $P_j \leftarrow P_i$ and $P_j \leftarrow P_k$. This results in two cross-path colliders in $P_j$. Now there are two possibilities, (a) they are both shielded which creates a directed path from X to $P_k$ and we recurse like before, or (b) at least one cross-path collider is unshielded and resulting in the sought after v-structure.

Since there are only k cross-path colliders, case 1, 2, and 3a reduce the number of colliders towards zero. If there are no cross-path colliders left, there is a directed path from X to $Prime(X)$ which contradicts our assumption that $\boldsymbol{Fb}(X) = \emptyset$. Therefore, eventually we must encounter case 3b and that proves it one way.

⇐ Assume all undirected paths between X and $Prime(X)$ have such a v-structure. We prove by contradiction that there does not exist a directed path from X to $Prime(X)$. Assume that $\boldsymbol{Fb}(X) \neq \emptyset$ and so there must be a path $P = X \rightarrow P_1 \rightarrow \ldots \rightarrow Prime(X)$, and assume it contains m such v-structures. Now let $P_i \rightarrow P_j \leftarrow P_k$ be the first v-structure (such that $j$ is the smallest). We consider:

1. $i > j$: There is a path $P_j \rightarrow \ldots \rightarrow P_i$ and also an edge $P_i \rightarrow P_j$ resulting in a cycle which is a contradiction.

2. $k > j$: Analogous to the first case.

3. $i, k < j$: Without loss of generality, assume that there is a path $X \rightarrow \ldots \rightarrow P_i \rightarrow \ldots \rightarrow P_k \rightarrow \ldots \rightarrow P_j \rightarrow$, and edges $P_i \rightarrow P_j$ and $P_k \rightarrow P_j$. So there is a directed path from X to $P_j$ without a v-structure and we recurse to find the first v-structure after $P_j$.

Since there are only m cross-path colliders, eventually there will be a path with no colliders left. Since this path contains no v-structures, it contradicts the fact that all paths must have a v-structure and, therefore, $\boldsymbol{Fb}(X) = \emptyset$. □